\documentclass{llncs}
\usepackage[numbers,sort&compress,sectionbib]{natbib}
\usepackage{amsmath,graphicx}
\usepackage{amsmath,graphicx}
\usepackage{acro}
\usepackage{tabularx}
\graphicspath{{Figure/}}
\usepackage{color}
\usepackage{multirow}
\usepackage{siunitx}
\usepackage{booktabs} 
\usepackage[font={small}]{caption, subfig}

\usepackage{epstopdf}
\usepackage{ulem}
\usepackage{arydshln}

\newcolumntype{Y}{>{\centering\arraybackslash}X}

\newcommand{\etal}{\textit{et al.}}

\newcommand{\Fig}{Fig.}

\definecolor{purple}{rgb}{0.6, 0, 0.6}
\newcommand{\raul}[1]{\textcolor{black}{#1}}
\newcommand{\dg}[1]{\textcolor{black}{#1}}
\usepackage{tikz}

\AtBeginDocument{}
\usepackage{amssymb}

\DeclareAcronym{CTV}{
short=CTV,
long= clinical target volume
}

\DeclareAcronym{GTV}{
short=GTV,
long= gross tumor volumec
}

\DeclareAcronym{RTCT}{
short=RTCT,
long= radiotherapy computed tomography
}

\DeclareAcronym{PET}{
short=PET,
long= positron emission tomography
}

\DeclareAcronym{RT}{
short=RT,
long= radiotherapy 
}

\DeclareAcronym{CAD}{
short=CAD,
long= computer aided diagnosis
}

\DeclareAcronym{PHNN}{
short=PHNN,
long= progressive holistically nested network
}

\DeclareAcronym{LN}{
short=LN,
long= lymph node
}

\DeclareAcronym{OSLN}{
short=GTV$_{LN}$,
long= lymph node gross tumor volume
}

\DeclareAcronym{SOTA}{
short=SOTA,
long= state-of-the-art
}

\DeclareAcronym{OAR}{
short=OAR,
long= organ at risk,
long-plural-form=organs at risk
}

\DeclareAcronym{SDT}{
short=DM,
long= distance transform map
}

\DeclareAcronym{CNN}{
short=CNN,
long= convolutional neural network
}

\DeclareAcronym{GNN}{
short=GNN,
long= graph neural network
}

\DeclareAcronym{CT}{
short=CT,
long= computed tomography
}

\DeclareAcronym{FP}{
short=FP,
long= false positive
}

\DeclareAcronym{DS}{
short=DS,
long= Dice score
}

\DeclareAcronym{ASD}{
short=ASD,
long= average surface distance
}

\DeclareAcronym{HD}{
short=HD,
long= Hausdorff distance
}

\DeclareAcronym{FCN}{
short=FCN,
long= fully convolutional network
}
\DeclareAcronym{VOI}{
short=VOI,
long= volume of interest
}

\DeclareAcronym{MLP}{
short=MLP,
long= multi-layer perceptron
}

\title{Lymph Node Gross Tumor Volume Detection in Oncology Imaging via Relationship Learning Using Graph Neural Network}

\author{Chun-Hung Chao\textsuperscript{1} \and Zhuotun Zhu$^\dag$\textsuperscript{2,3} \and Dazhou Guo$^\dag$\textsuperscript{2} \and Ke Yan$^\dag$\textsuperscript{2} \and Tsung-Ying Ho\textsuperscript{4} \and Jinzheng Cai\textsuperscript{2} \and Adam P. Harrison\textsuperscript{2} \and Xianghua Ye\textsuperscript{5} \and Jing Xiao\textsuperscript{6} \and Alan Yuille\textsuperscript{3} \and Min Sun\textsuperscript{1} \and Le Lu\textsuperscript{2} \and Dakai Jin\textsuperscript{2} }

\institute{\textsuperscript{1}National Tsing Hua University, Hsinchu City, Taiwan, ROC \\ \textsuperscript{2}PAII Inc., Bethesda, MD, USA \\ \textsuperscript{3} Johns Hopkins University, Baltimore, MD, USA \\  \textsuperscript{4}Chang Gung Memorial Hospital, Linkou, Taiwan, ROC \\
\textsuperscript{5}The First Affiliated Hospital Zhejiang University, Hangzhou, China \\
\textsuperscript{6}Ping An Technology, Shenzhen, China \\
 }

\newcommand\blfootnote[1]{%
  \begingroup
  \renewcommand\thefootnote{}\footnote{#1}%
  \addtocounter{footnote}{-1}%
  \endgroup
}

\begin{document}
\setlength{\abovecaptionskip}{1ex}
\setlength{\belowcaptionskip}{1ex}
\setlength{\floatsep}{1ex}

\maketitle

\begin{abstract}
Determining the spread of \ac{OSLN} is essential in defining the respective resection or irradiating regions for the downstream workflows of surgical resection and radiotherapy for many cancers. Different from the more common enlarged \ac{LN}, \ac{OSLN} also includes smaller ones if associated with high positron emission tomography signals and/or any metastasis signs in CT. This is a daunting task. In this work, we propose a unified LN appearance and inter-\ac{LN} relationship learning framework to detect the true \ac{OSLN}. This is motivated by the prior clinical knowledge that \acp{LN} form a connected lymphatic system, and the spread of cancer cells among \acp{LN} often follows certain pathways. Specifically, we first utilize a 3D convolutional neural network with ROI-pooling to extract the \ac{OSLN}'s instance-wise appearance features. Next, we introduce a graph neural network to further model the inter-\ac{LN} relationships where the global LN-tumor spatial priors are included in the learning process. This leads to an end-to-end trainable network to detect by classifying \ac{OSLN}. We operate our model on a set of \ac{OSLN} candidates generated by a preliminary 1st-stage method, which has a sensitivity of $>85\%$ at the cost of high \ac{FP} ($>15$ FPs per patient). We validate our approach on a radiotherapy dataset with 142 paired PET/RTCT scans containing the chest and upper abdominal body parts. The proposed method significantly improves over the \ac{SOTA} \ac{LN} classification method by $5.5\%$ and $13.1\%$ in F1 score and the averaged sensitivity value at $2, 3, 4, 6$ \acp{FP} per patient, respectively.

\end{abstract}


\blfootnote{$^\dag$ equal contribution.}

\acresetall

\section{Introduction}
\label{sec:intro}

Quantitative \ac{LN} analysis is an important clinical task for cancer staging and identifying the proper treatment areas in radiotherapy. The revised RECIST guideline~\cite{eisenhauer2009new} recommends measuring enlarged \acp{LN} (if short axis $>10$ mm) for the purpose of tumor burden assessment. However, in cancer treatment, like radiotherapy or surgery, besides the primary tumor, all metastasis-suspicious \acp{LN} are also required to be treated. This includes the enlarged \acp{LN}, as well as other smaller ones that are associated with high \ac{PET} signals and/or other metastasis signs in CT. This broader category is referred as \ac{OSLN} in radiotherapy treatment. Accurate identification of \ac{OSLN} is essential for the delineation of clinical target volume in radiotherapy~\cite{jin2019deep}, where missing small but involved \ac{OSLN}, will lead to undesired under-treatment~\cite{National2020}.

\begin{figure}[t]
\centering
\includegraphics[scale=.29]{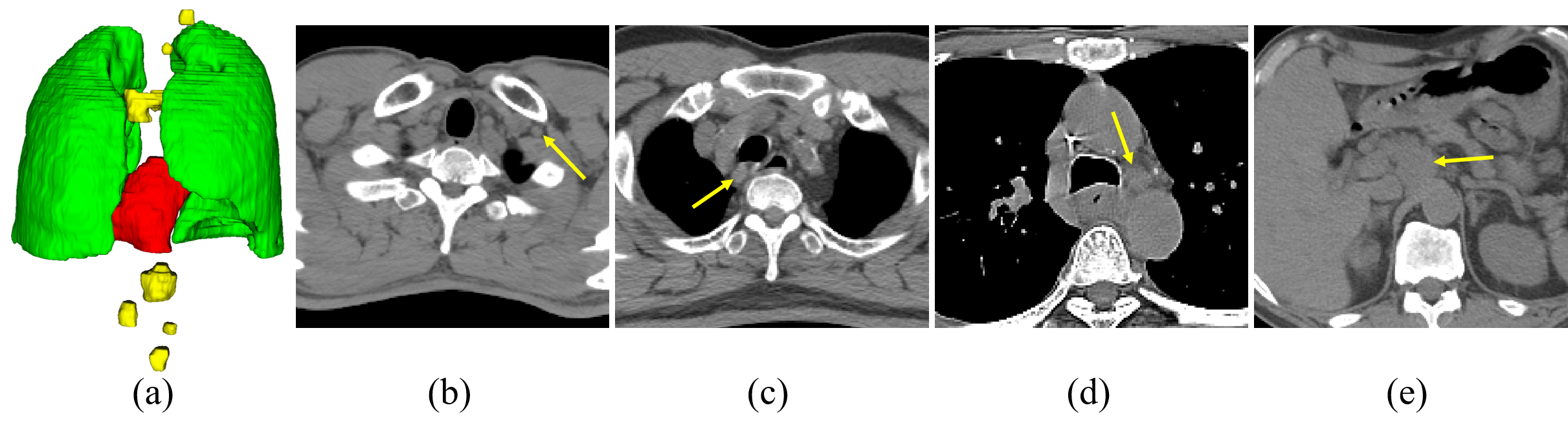}
\caption{(a) A 3D rendering of an esophageal tumor (red), lung (green) and the \acp{LN} identified by the oncologist (yellow). (b-e) The axial view of the \ac{OSLN} spanning from the lower neck to the upper abdomen region. Note that the \ac{OSLN} includes the enlarged \ac{LN}, as well as smaller but suspected metastasized ones. }
\label{Fig:motivation}
\end{figure}

Identifying \ac{OSLN} could be a very difficult and time-consuming task, even for experienced radiation oncologists. It requires a sophisticated subjective reasoning process that leads to high inter-observer variabilities~\cite{goel2017clinical}. Beside the difficulties already met by detecting enlarged \acp{LN}, \acp{LN} can exhibit low contrast with surrounding tissues and can be easily confused with other vessel or muscle structures. \ac{OSLN} detection has its unique challenges: (1) \ac{OSLN} discovery is often performed using radiotherapy CT (RTCT), which, unlike diagnostic CT, is not contrast-enhanced. (2) The size and shape of \ac{OSLN} vary considerably, and their locations have an implicit relation with the primary tumor~\cite{akiyama1994radical}. See Fig.~\ref{Fig:motivation} for an illustration of \ac{OSLN}. While many previous work developed automatic detection/identification methods for enlarged \acp{LN} using contrast-enhanced CT ~\cite{barbu2011automatic,bouget2019semantic,feulner2013lymph,roth2015improving,yan2019mulan}, not much work has attempted on the \ac{OSLN} identification task. These previous work all focus on characterizing instance-wise \ac{LN} features. Note that, similar to~\cite{seff2015leveraging,roth2015improving,roth2014new}, our task assumes there is a set of \ac{OSLN} candidates computed from an existing detection system with high sensitivities, but low precision so that we target on effectively reducing the \ac{FP} \ac{OSLN}. As shown in our experiments, applying the previous state-of-the-art \ac{LN} identification by classification method~\cite{roth2015improving} leads to a markedly inferior performance.

Unlike prior work that only assess or identify individual \ac{LN} separately and independently, we perform a \emph{study-wise} analysis that incorporates the inter-\ac{LN} and \ac{LN}-primary tumor relationships. This is motivated by the fact that our lymphatic system is a connected network of \acp{LN}, and tumorous cells often follow certain pathways to spread between the \acp{LN}~\cite{akiyama1994radical}. To achieve this, we propose to use \acp{GNN} to model this inter-\acp{LN} relationship.  Specifically, we first train a 3D \ac{CNN} to extract \ac{OSLN} instance-wise appearance features from CT. Then, we compute the 3D distances and angles for each \ac{OSLN} with respect to the primary tumor, which serves as the spatial prior of each \ac{OSLN} instance. Using the instance-wise appearance features and spatial priors to create node representations, a \ac{GNN} is built, where relationships between \ac{OSLN} are patient adaptive via a learnable features fusion function. This allows the \ac{GNN} to automatically learn the \ac{OSLN}-candidate connection strengths to help distinguishing between true and false \ac{OSLN}. The whole \ac{CNN} and \ac{GNN} framework is end-to-end trainable, allowing \ac{GNN} to guide the appearance feature extraction in the \ac{CNN}.  Moreover, PET imaging is included as an additional input channel to the \ac{CNN} model to provide CT imaging with complementary oncology information, which is demonstratively helpful for \ac{OSLN} identification task~\cite{goel2017clinical}. \dg{We evaluate on a dataset of $142$ esophageal cancer patients, as the largest dataset to date. The proposed} method significantly improves over the \ac{SOTA} \ac{LN} detection/classification method ~\cite{roth2015improving,roth2014new} by \dg{$5.5\%$ and $13.1\%$ in F1 scores and the averaged sensitivity at $2, 3, 4, 6$ \acp{FP} per patient, respectively. }

\section{Methods}
\label{sec:method}
\begin{figure}[t]
\centering
\includegraphics[width=0.99\textwidth]{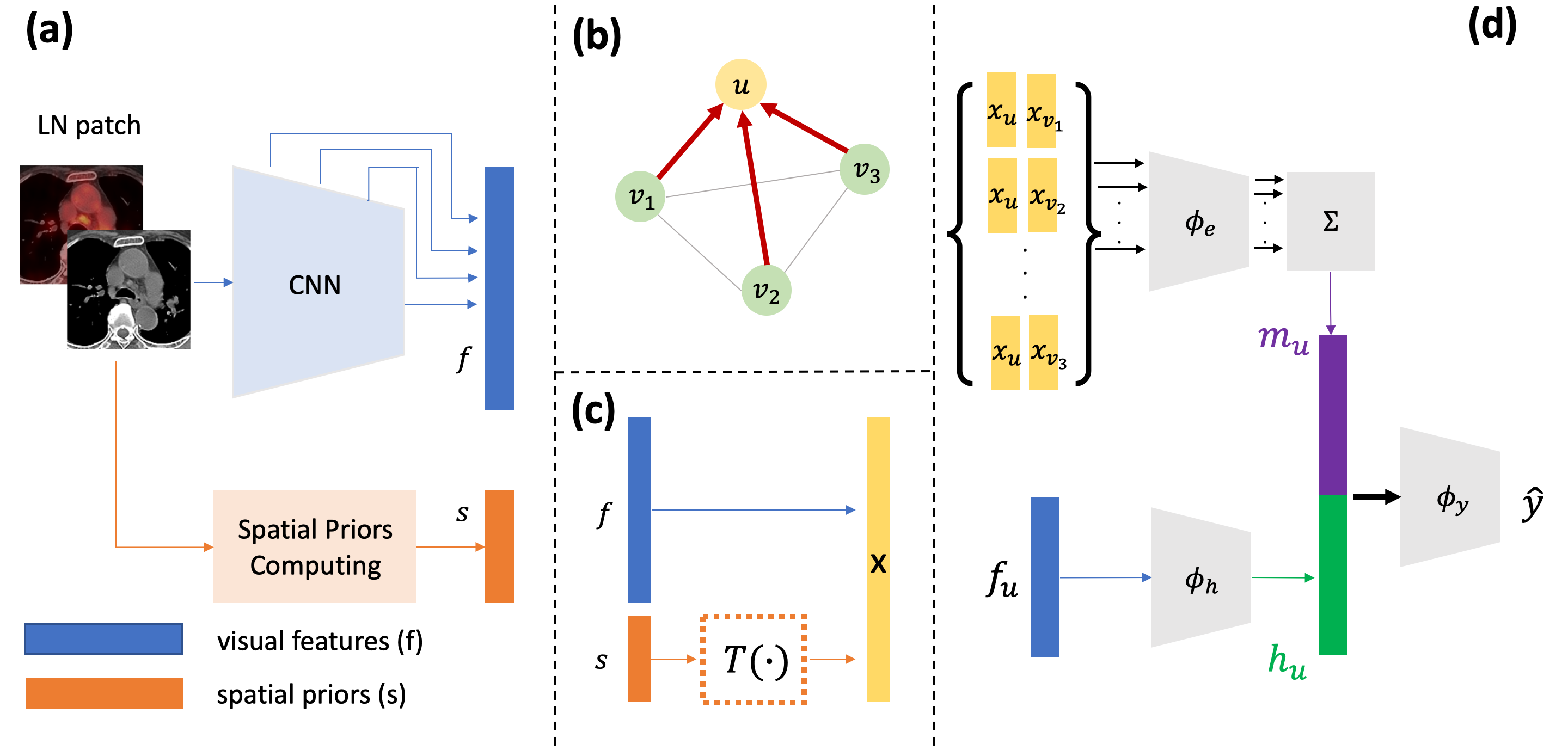}
\caption{\textbf{System overview.} (a) For each \ac{OSLN} candidate, the model extracts \acs{CNN} appearance features, $f$, and the spatial priors, $s$, for each candidate to create node representations, $x$. (b) We treat each \ac{OSLN} candidate as a node in the graph and use proposed \acs{GNN} to exchange information. Here we take node $u$ as an example for a target node and the red arrow denotes message passing from other nodes. (c) and (d) show how we obtain the node representations and node latent features $h$, respectively. (d) also depicts how the aggregated passed message, $m$, and latent features are used to identify \ac{OSLN}.}
\label{Fig:workflow}
\end{figure}

Our \ac{OSLN} approach combines 3D \ac{CNN} and \ac{GNN} networks. \Fig~\ref{Fig:workflow} depicts an overview of our method, which consists of three modularized components: (1) a 3D CNN classifier to extract per \ac{OSLN} candidate instance-wise visual features; (2) spatial prior computation, including the \ac{LN}-to-tumor distance and angle calculations; (3) a \ac{GNN} that learns the inter-\ac{LN} candidate relationship using the global \ac{OSLN} spatial priors and their instance-wise CNN features.

\subsection{3D CNN-based Appearance Learning}
\label{sec:local_module}
We first train a \ac{CNN} to extract the \ac{OSLN} instance-wise visual features from CT and PET imaging. We adopt a multi-scale 3D CNN model with a ROI-GAP layer~\cite{girshick2015fast} that accepts paired CT/PET image patches. The features generated by each convolutional block separately pass through a 3D ROI-GAP layer and a fully connected layer to form a $256$-dimensional vector, which are then concatenated together to generate a multi-scale local representation for each \ac{OSLN} candidate. Since we use four CNN blocks, this leads to a $4\times256=1024$-dimensional feature vector, which is denoted as $f$.

\subsection{Spatial Prior Computation}
\label{sec:spatial_prior}

In addition to the appearance features, spatial priors can provide crucial information for generating the \ac{GNN} node representation. We compute each normalized 3D spatial coordinates and also 3D distance and angles to the primary tumor. First, to calculate the normalized 3D spatial coordinates  we use each patient's lung mask range in the x and y axes to normalize the \ac{OSLN} candidates' $x$ and $y$ locations. Specifically, we adopt and reimplement a pretrained a 2D \ac{PHNN}~\cite{harrison2017progressive} to segment the lung, which can perform robustly on pathological lungs. Then we calculate the minimum and maximum coordinates of the lung mask and use these as cutoffs to normalize all candidate $x$ and $y$ location to the range of $[0, 1]$. For the $z$ dimension, we train a fully unsupervised body-part regressor~\cite{yan2018unsupervised} using our dataset to provide the normalized $z$ location for each \ac{OSLN} candidate instance.

To compute the distance and angles from any candidate to the primary tumor, we first compute the 3D distance transform from the tumor segmentation mask $\mathcal{O}$. Let $\Gamma(\mathcal{O})$ be a function that computes all boundary voxels of the tumor mask. The \ac{SDT} value at a voxel $p$ with respect to $\mathcal{O}$ is computed as
\begin{align}
\text{DM}_{\Gamma(\mathcal{O}_i)}(p) = \left \{
\begin{array}{rcl}
\underset{q\in \Gamma(\mathcal{O}_i)}{\min} d(p,q)  & \quad {\text{if} \quad p\notin \mathcal{O}_i}\\
0  & \quad {\text{if} \quad p\in \mathcal{O}_i }
\end{array} \right. \mathrm{,}
\end{align}
where $d(p,q)$ is a distance measure from $p$ to $q$. We use the Euclidean distance metric and apply Maurer \etal{}'s efficient algorithm~\cite{maurer2003linear} to compute the \acp{SDT}. For calculating the angles of each \ac{OSLN} candidate with respect to the tumor, we extract the center point or centerline~\cite{jin2016robust} of the tumor and then calculate the elevation ($\theta$) and azimuth ($\phi$) with respect to the candidate center. We denote all these spatial priors as $s$.

\subsection{GNN-based Iner-LN Relationship Learning} 
\label{sec:global_module}

{\bf Inter-LN Graph Formulation}
\raul{With the local visual features obtained from the CNN model and the spatial priors calculated in Sec~\ref{sec:spatial_prior}, we use them to create a node representation, $x$, for each \ac{OSLN} candidate. Because of divergence  in construction between the visual features and spatial priors, we use a learnable function, e.g., a \ac{MLP}, $T$ to transform the latter to align them better with the CNN visual features: 
\begin{equation}
    \begin{aligned}
        s^{\prime} &= T(s) \mathrm{,}\\
        x &= (f \parallel s^{\prime}) \mathrm{,}
    \end{aligned}
\end{equation}
\\
where $\parallel$ denotes concatenation. In addition to the node representation, each node has its own latent feature $h$ produced by a learnable function $\phi_h$. Since the latent feature is considered a local feature that does not  take part in information exchange, we only use the visual features here:
\begin{equation}
\label{eqn:latent}
    h = \phi_h(f) \mathrm{.}
\end{equation}
}

{\bf Graph Message Passing Neural Network}
Message passing neural networks~\cite{gilmer2017neural} are a widely used basis for \acp{GNN}, such as gated graph sequential networks~\cite{li2015gated}, graph attention networks~\cite{velivckovic2017graph}, and dynamic graph message passing networks~\cite{zhang2019dynamic}. Given the target node $u$ and its neighboring nodes $\mathcal{N}(u)$, the key idea is to collect information from all of the neighboring nodes, which are usually their node representations and extract the essential information:
\begin{equation}
\label{eqn:orig_update}
    m_u = \sum_{v \in \mathcal{N}(u)}{\alpha_{uv} \cdot \phi_e(x_v)} \mathrm{,}
\end{equation}
where $m_u$ denotes the message passed to the target node and a function $\phi_e$ is used to aggregate information from the node neighborhood. Note that all extracted information is weighted by the distance $\alpha\in [0,1]$ between the neighboring and the target nodes. 

The target node $u$ then updates its own latent feature $h_u$ with the message $m_u$ using an updating function $\phi_u$:
\begin{equation}
\label{eqn:gnnupdate}
    h_u^{\prime} = \phi_u(h_u, m_u) \mathrm{.}
\end{equation}
Common choices for $\phi_u$ are the linear, gated attention or fully connected layers.

{\bf Inter-LN Relationship Modeling}
\raul{The above graph message passing network weights the latent feature vector $h$ by a scalar $\alpha$, as in \eqref{eqn:orig_update}. However, a more adaptive approach can allow for a more powerful model of the inter-nodal relationship. Hence, we propose to use the representations of both source node $x_v$ and target node $x_u$ to implicitly model the inter-nodal relationship and generate more informative messages to pass to the target. More formally, we use a learnable function, $\mathcal{G}$, which can fuse the feature vectors from pairs of nodes:
\begin{equation}
    m_u = \sum_{v \in \mathcal{N}(u)} \phi_{e}(\mathcal{G}(x_u, x_v)) \mathrm{.}
\end{equation}
Note that we consider our graph as a fully connected graph with self-connections, thus all the nodes belong to the neighborhood $\mathcal{N}(u)$ of any particular node $u$. Once all the nodes have their latent features updated, the predictions of each node are made based on their $h_u^{\prime}$, which aggregates information all nodes in the graph:
\begin{equation}
    \hat{y} = \phi_y(h_u^{\prime}) \mathrm{.}
\end{equation}
}

\section{Experiments and Results}
\noindent\textbf{Dataset} We collected a dataset of $142$ esophageal patients who underwent radiotherapy treatment. In total, there are $651$ \ac{OSLN} in the mediastinum or upper abdomen region that were identified by an oncologist. Each patient has a non-contrast RTCT scan and a PET scan that has been registered to the RTCT using the method of \cite{jin2019accurate}. We randomly split patients into $60\%$, $10\%$, $30\%$ for training, validation and testing, respectively. 

\noindent\textbf{\ac{OSLN} candidates generation}
We first use an in-house \ac{OSLN} CAD system to generate the \ac{OSLN} candidates that will be used in this work~\cite{zhu2020detecting}. The CAD system achieves $85\%$ sensitivity with a large number of \ac{FP} detections ($>15$ FPs per patient). This leads to $>2000$ \acp{FP} that serve as negative \ac{OSLN} candidates. For the ease of comparison, similar to~\cite{roth2015improving}, we also include the ground-truth \ac{OSLN} in the set of true \ac{OSLN} candidates for training. This ensures $100\%$ sensitivity at the \ac{OSLN} candidate generation step. 

\noindent\textbf{Image preprocessing and implementation details} We resample the RTCT and registered PET images to a consistent spatial resolution of $1.0\times1.0\times1.0$ mm. The 3D training patch is generated by cropping a $48 \times 48 \times 48$ sub-volume centered around each \ac{OSLN} candidate. If the size of the \ac{OSLN} is larger than $48 \times 48 \times 48$, we resize the sub-volume so that it contains at least an $8$-voxel margin of the background along each dimension to ensure sufficient background context. For the training sample generation for the 2.5D LN classification method~\cite{roth2015improving}, we adopt the preprocessing method described in that paper. Our 3D CNN is trained using the Adam~\cite{kingma2014adam} optimizer with a learning rate of $0.0001$ and batch size of $32$ for $10$ epochs. 

\raul{For the GNN part, the parameters of the first layer in the CNN are loaded from the CNN baseline model reported in Table~\ref{tab:quantres}. The feature aligning function $T$ for spatial priors can be either a simple repeat function or a \ac{MLP}. We opt for the latter approach, choosing an output dimension size of $640$. We also use \acp{MLP} for $\phi_e$ and $\phi_h$ which reduce the input dimension from $3328$ and $1024$, respectively, to $256$. 
The entire \ac{GNN} architecture is trained with Adam optimizer whose learning rate is set to $0.0001$. Since the numbers of \ac{OSLN} candidates vary from patient to patient, the nodes in GNN changed dynamically. To accommodate different graph sizes, we set batch size as 1 and adopted gradient accumulation strategy which is equivalent to using batch size 8 during training.}

\noindent\textbf{Comparison setup and evaluation metrics} We compare against the \ac{SOTA} instance-wise LN classification method~\cite{roth2015improving}, which uses a random view aggregation in a 2.5D fashion to learn the LN local representations. We also compare against a CNN-based classifier (both 2.5D and our 3D input) under various input settings: CT alone, CT+PET. We also compared against the GNN with the weight in the adjacency matrix ($\alpha_{uv}$ in \eqref{eqn:orig_update}) regressed by an additional module~\cite{velivckovic2017graph}, denoted as {\bf CNN+GNN$_b$}. Our method is denoted as {\bf CNN+GNN$_p$}.

To evaluate performance, we compute the free response operating characteristic (FROC), which measures the recall against different numbers of FPs allowed per patient. We also report the average sensitivity (mFROC) at $2$, $3$, $4$, $6$ FPs per patient.

\begin{table}[]
\caption{Quantitative results of our proposed methods with the comparison to other setups and the previous satate-of-the-art. Note, we use ResNet18 as the CNN backbone. The $\text{GNN}_b$ and $\text{GNN}_p$ denote the baseline and proposed \acp{GNN}, respectively. S.P. denotes the spatial prior features of \ac{OSLN}.}
\centering
\begin{tabular}{l|cc|ccc|rlr}
\hline
Method               & 2.5D       & 3D         & CT         & PET        & S.P.      & F1 &  & mFROC \\ \hline
Roth et. al~\cite{roth2015improving}                  & \checkmark &            & \checkmark &            &            & 0.483    &  & 0.537        \\
Roth et. al~\cite{roth2015improving}                  & \checkmark &            & \checkmark & \checkmark &            & 0.498    &  & 0.561        \\ \hline
CNN                  &            & \checkmark & \checkmark &            &            & 0.493    &  & 0.568        \\
CNN                  &            & \checkmark & \checkmark &            & \checkmark & 0.516    &  & 0.602        \\ \hdashline
CNN                  &            & \checkmark & \checkmark & \checkmark &            & 0.517    &  & 0.595        \\
CNN                  &            & \checkmark & \checkmark & \checkmark & \checkmark & 0.505    &  & 0.597        \\ \hline
CNN + $\text{GNN}_b$ &            & \checkmark & \checkmark &            & \checkmark & 0.500    &  & 0.593        \\
CNN + $\text{GNN}_b$ &            & \checkmark & \checkmark & \checkmark & \checkmark & 0.521    &  & 0.631        \\ \hdashline
CNN + $\text{GNN}_p$ &            & \checkmark & \checkmark &            & \checkmark & 0.504    &  & 0.631        \\
CNN + $\text{GNN}_p$ &            & \checkmark & \checkmark & \checkmark & \checkmark & 0.538    &  & 0.668        \\ \hline
\end{tabular}
\label{tab:quantres}
\end{table}

{\bf Quantitative Results and Discussions}
Table~\ref{tab:quantres} outlines the quantitative comparisons of different model setups and choices. First, several conclusions can be drawn to validate the effectiveness of our CNN-based classification method.  (1) The SOTA instance-wise 2.5D LN classification method~\cite{roth2015improving} exhibits markedly decreased performance as compared to our 3D CNN in both CT and CT+PET input settings, {\it{e.g.}}, \cite{roth2015improving} has a mFROC of $0.561$ with CT+PET inputs as compared to $0.595$ achieved by our 3D CNN model with CT+PET inputs. This demonstrates that the direct 3D convolution is at least as effective as the pseudo 3D method~\cite{roth2015improving} for the LN identification problem.  (2) PET modality plays an important role in the \ac{OSLN} detection, since both 2.5D and 3D CNN gain consistent mFROC improvement after adding PET as an additional input channel. (3) Spatial prior features are useful for the CT based CNN classifier, however, it does not add value when the CNN is already trained with CT+PET. This may be due to the fact that the instance-wise \ac{OSLN} features extracted from CT+PET reached the saturated classification capacity, however, the spatial priors provide useful information if features are extracted from CT alone.   

\begin{figure}[t!]
\centering
\includegraphics[width=0.96\textwidth]{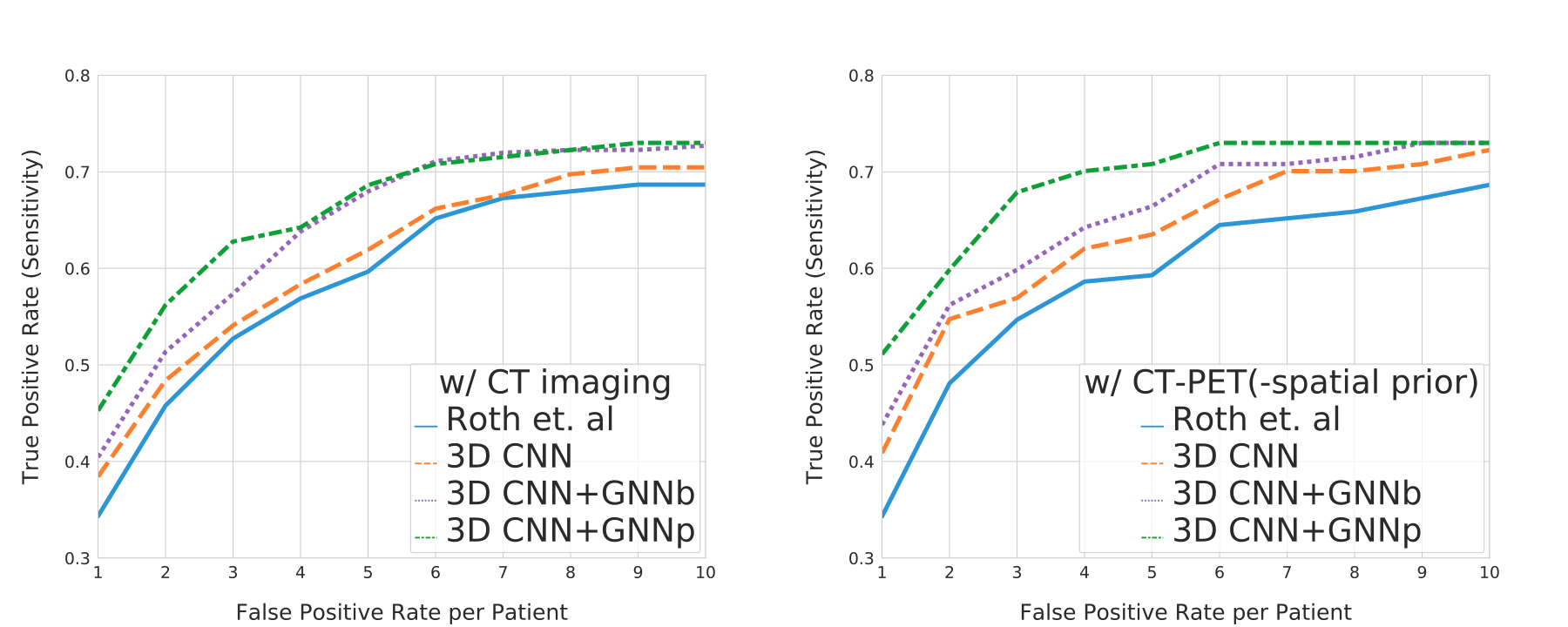}
\caption{Free-response receiver operating characteristic (FROC) curves of different \ac{OSLN} classification methods under the setting of {\bf CT} ({\bf Left}) imaging or {\bf CT+PET} ({\bf Right}) imaging. The proposed method (GNN$_p$) outperforms the baseline 3D CNN and the node-level feature fusion GNN (GNN$_b$) by some marked margins under both imaging settings.}
\label{fig:froc}
\end{figure}

\begin{figure}[t!]
\centering
\includegraphics[width=0.96\textwidth]{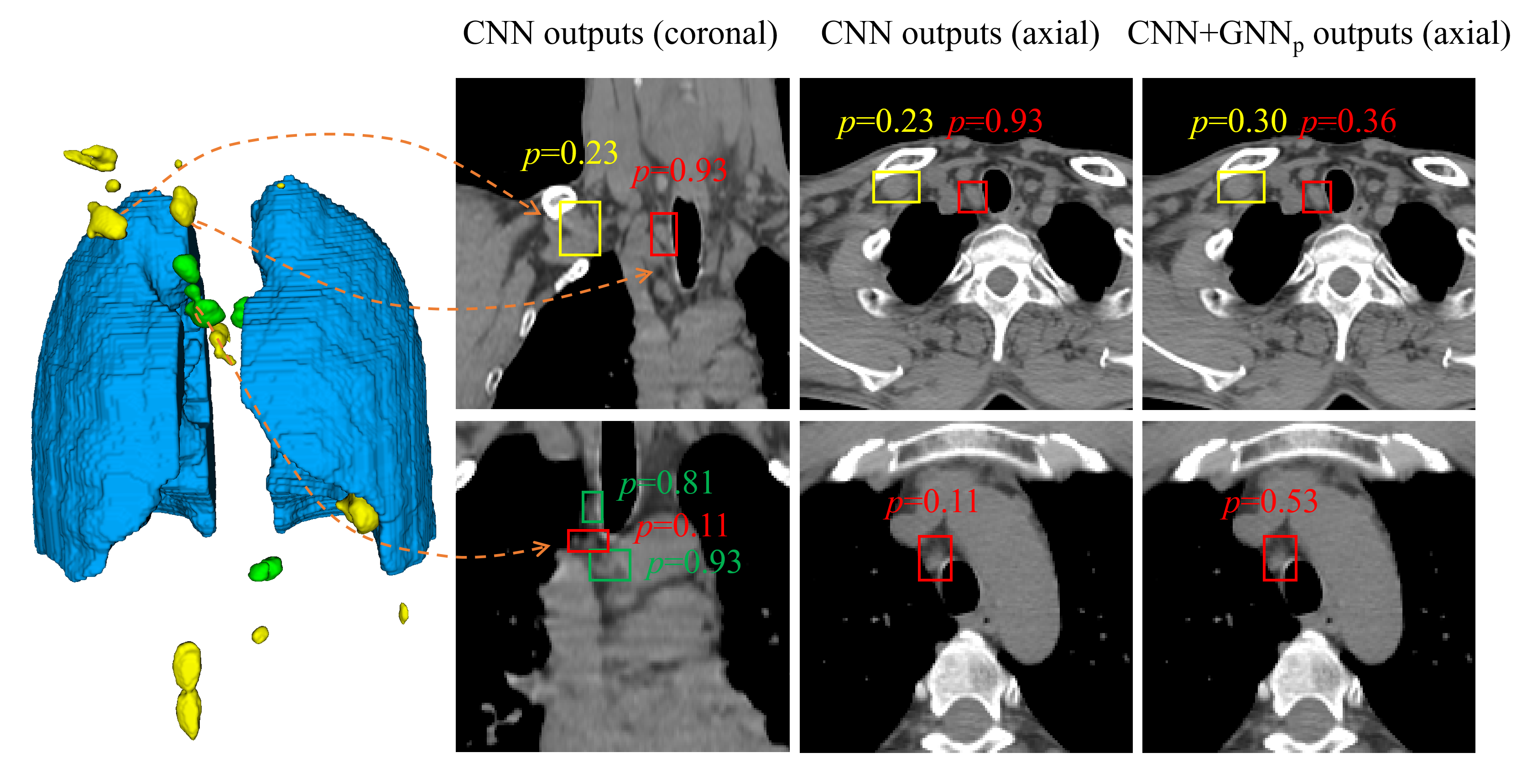}
\caption{A quality example illustrating the classification probabilities computed by the 3D CNN based LN instance-wise classifier and by our proposed CNN+GNN$_p$ classifier. Left is the 3D rendering of the \ac{OSLN} candidates with yellow as FPs from an in house CAD system~\cite{zhu2020detecting} and green as the true \ac{OSLN}. The top row shows an example where a FP exists at the mediastinum region near the shoulder by the CNN classifier (shown in red with probability of 0.93). Due to the fact that a nearby LN candidate has a low probability of $0.23$ (indicate in yellow), and they share a similar location and appearance. This candidate is correctly identified as FP by our CNN+GNN$_p$ classifier. Similar situation is observed for the right bottom row, where a small LN (shown in red) has misclassified by the CNN model, but is able to correctly identified by our model due to the fact that there are two nearby strong true candidates (indicated in green) and their appearance is similar. }
\label{fig:quality}
\end{figure}

When further considering the inter-LN relation captured by the GNN model, the \ac{OSLN} classification performance is consistently boosted. For example, under the CT+PET input with the spatial prior setting, both the CNN+GNN$_b$ and CNN+GNN$_p$ outperform the corresponding 3D CNN model by a large margin, {\it{i.e.}}, $3.4\%$ and $6.9\%$ in mFROC metrics, respectively. This validates our observation and intuition that the LNs form a connected network and cancers often follow certain pathways spreading to the LNs. Hence, learning the inter-LN relationship should be beneficial to distinguish the metastasis-suspicious \ac{OSLN}. Moreover, our proposed element-level feature fusion method in the GNN achieves the highest classification performance under both CT and CT+PET input settings as compared to a competing node-level feature fusion method~\cite{velivckovic2017graph}. FROC curves are compared in Fig.~\ref{fig:froc} for the CNN$+$GNN methods and the 3D CNN method under different imaging settings. It can be seen that at an operating point of 3 FP/patient, our CNN$+$GNN$_p$ improve the sensitivity by $7.3\%$ ($55.5\%$ to $62.8\%$) and $11.0\%$ ($56.9\%$ to $67.9\%$) over the 3D CNN model using the CT and
CT+PET imaging, respectively. This further demonstrates the value of the inter-LN relationship learning in the \ac{OSLN} classification task. Some qualitative examples are illustrated in Fig.~\ref{fig:quality}. 

\section{Conclusion}
\label{sec:conclusion}
In this paper, we propose a joint 3D deep learning framework on modeling both LN appearance and inter-\ac{LN} relationship for effective identification of \ac{OSLN}. The inter-\ac{LN} relationship is expected to capture the prior spatial structure of the connected lymphatic system where the spread of cancer cells must follow certain pathways. We combine the process of two 3D CNN (appearance encoding) and GNN (iner-LN relationship learning) networks, which still leads to an end-to-end trainable network. We perform our unified CNN-GNN identification by classification model on a set of \ac{OSLN} candidates that are generated by a preliminary 1st-stage method with very high detection sensitivity, but at a lower precision level. We validate our approach on a esophageal radiotherapy dataset of 142 patients and total 651 \ac{OSLN}. Our quantitative performance demonstrates significant improvements over previous state-of-the-art LN classification work.

\bibliographystyle{splncs04}
\bibliography{refs}
\end{document}